\documentclass[]{spie}  

\usepackage{amsmath,amsfonts,amssymb}
\usepackage{graphicx}
\usepackage{bbding}
\usepackage[colorlinks=true, allcolors=blue]{hyperref}

\title{Anatomically-Informed Data Augmentation for functional MRI with Applications to Deep Learning}

\author{Kevin P. Nguyen \href{mailto:kevin3.nguyen@utsouthwestern.edu}{\Envelope}}
\author{Cherise Chin Fatt}
\author{Alex Treacher}
\author{Cooper Mellema}
\author{Madhukar H. Trivedi}
\author{Albert Montillo}
\affil{UT Southwestern Medical Center, Dallas, TX, USA}

\begin{document} 
\maketitle


\section{ABSTRACT}
The application of deep learning to build accurate predictive models from functional neuroimaging data is often hindered by limited dataset sizes. Though data augmentation can help mitigate such training obstacles, most data augmentation methods have been developed for natural images as in computer vision tasks such as CIFAR, not for medical images. This work helps to fills in this gap by proposing a method for generating new functional Magnetic Resonance Images (fMRI) with realistic brain morphology. This method is tested on a challenging task of predicting antidepressant treatment response from pre-treatment task-based fMRI and demonstrates a 26\% improvement in performance in predicting response using augmented images. This improvement compares favorably to state-of-the-art augmentation methods for natural images. Through an ablative test, augmentation is also shown to substantively improve performance when applied before hyperparameter optimization. These results suggest the optimal order of operations and support the role of data augmentation method for improving predictive performance in tasks using fMRI.

\section{INTRODUCTION}
\label{sec:intro}  

Neural networks have proved to be powerful modeling tools for many medical imaging problems, such as predicting neurological and psychiatric diagnoses and prognoses from brain MRI. However, the training of neural networks for these problems is frequently hindered by small dataset sizes, making it challenging to produce high-performing, generalizable models. Data augmentation, which synthesizes additional data samples from real data, has improved performance in many non-medical deep learning problems such as natural image classification. For example, recent methods such as AutoAugment \cite{Cubuk.5242018} and Population-Based Augmentation \cite{Ho.5142019} have improved classification error rate on highly studied datasets such as CIFAR, SVHN, and MNIST by up to 1.5\% (a 12\% relative improvement from previous state-of-the-art). On a reduced CIFAR dataset, the performance benefit of these AutoAugment was as high as 7\%, highlighting the importance of data augmentation in cases of limited dataset size. However, data augmentation techniques developed for natural images typically involve color transformation and geometric transformations such as flipping and shearing, which may not be suitable for brain images because they introduce transformations that do not yield realistic brain appearance and morphology. In other words, these operations can produce implausible brain images. For brain MRI, one method for augmenting structural MRI (sMRI), involving independent components analysis (ICA) and random loading matrices, has shown to improve the accuracy of schizophrenia vs. healthy control classification by 5\% (7-8\% relative improvement using augmentation) \cite{Ulloa.91720159202015, Castro.41520154182015}. No such method has been developed and validated for functional MRI (fMRI).

The contributions of this work are as follows. 1) A novel, coregistration-based fMRI data augmentation method is proposed, which synthesizes new realistic raw fMRI images. 2) The performance benefit of this augmentation is demonstrated on an antidepressant response prediction task, where the goal is to produce a pre-treatment predictor of clinical response to a commonly used antidepressant, sertraline. Since individual antidepressant response is highly variable, improving the accuracy of such a predictor would help reduce morbidity in Major Depressive Disorder (MDD) by aiding clinicians in identifying MDD patients most likely to benefit from sertraline. 3) Additionally, this work provides evidence that augmentation not only improves overall model performance but also enables the identification of better models during model hyperparameter optimization.

{\let\thefootnote\relax\footnote{{Accepted to SPIE Medical Imaging, February 2020.}}}

\section{METHODS}
\label{sec:methods}
\paragraph{Materials}
\label{subsec:materials}
Data from the Establishing Moderators and  Biosignatures of Antidepressant Response in Clinical Care (EMBARC) study \cite{Trivedi.2016}, a randomized controlled clinical trial, was used for the following experiments. This dataset contains 163 MDD subjects who underwent pre-treatment sMRI and task-based fMRI, then completed an 8-week treatment course with the antidepressant sertraline. The predictive task is to estimate the change in clinical severity between pre-treatment and week 8 of treatment from pre-treatment fMRI. This severity is measured by the Hamilton Rating Scale for Depression (HAMD) score. Demographics and pre-treatment clinical measurements, including psychiatric scales, comorbidities, and disease duration, are also added to the predictive models as covariates.

T1-weighted sMRI was acquired at 3T with the MPRAGE sequence; TE was 2.4 ms or 3.7 ms depending on study site, with dimensions of $256 \times 256 \times 176$, and an isotropic voxel size of 1 mm. BOLD fMRI was acquired using GE-EPI, a TR of 200 ms, dimensions of $64 \times 64 \times 39$, and voxel size of $3.2 \times 3.2 \times 3.1$ mm for 8 minutes. During fMRI acquisition, subjects completed a block-design number-guessing task that stimulates reward processing circuitry known to be altered in depressed individuals \cite{Greenberg.2015}.

\begin{figure} [ht]
\begin{center}
\includegraphics[width=\textwidth]{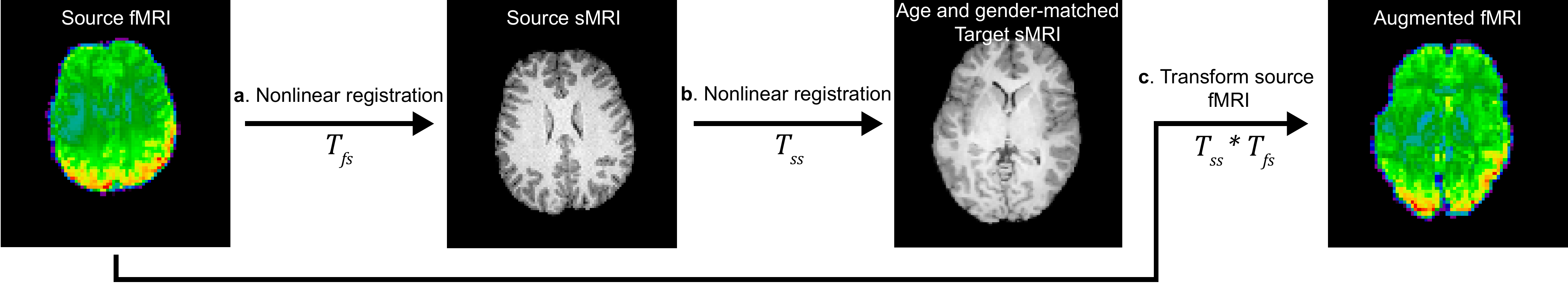}
\end{center}
\caption[example] 
{ \label{fig:augmentation} 
The proposed data augmentation method synthesizes a new fMRI image by performing a T1-based coregistration to another subject's brain in native space. \textbf{a.} The \textit{source} fMRI mean frame is registered to the \textit{source} sMRI. \textbf{b.} The \textit{source} sMRI is registered to an age- and gender-matched \textit{target} sMRI.  \textbf{c.} The combination of these transformations is  applied to transform the \textit{source} fMRI into a synthetic fMRI in \textit{target} space.}
\end{figure}

\paragraph{Data Augmentation}
\label{subsec:augmentation}
To generate anatomically-constrained synthetic fMRI images from real data, the proposed method employs a T1-based coregistration scheme to precisely resample a \textit{source} subject's original fMRI signal onto the brain anatomy of a \textit{target} subject in native space (\textbf{Fig. \ref{fig:augmentation}}). First, brain extraction is performed on the \textit{source} and \textit{target} subjects' \textbf{s}MRI using the ROBEX software. Brain extraction is then performed on the \textit{source} subject's mean \textbf{f}MRI volume using a combination of FSL BET and AFNI 3dAutomask tools. Next (\textbf{\ref{fig:augmentation}a}), the \textit{source} fMRI mean frame is coregistered to the source sMRI using the \texttt{antsRegistrationSyNQuick} routine in ANTS, which performs a sequence of multi-scale linear and nonlinear registrations. Then (\textbf{\ref{fig:augmentation}b}), the \textit{source} sMRI is coregistered to the \textit{target} sMRI using \texttt{antsRegistrationSyN} which has been shown to accurately coregister brain anatomy in sMRI across different subjects. Finally (\textbf{\ref{fig:augmentation}c}), the transformations from steps \textbf{a}) and \textbf{b}) are combined and applied to the \textit{source} fMRI. This creates a new image with the \textit{source}'s fMRI signal in the \textit{target}'s brain space. 

\paragraph{fMRI Preprocessing}
\label{subsec:preprocessing}
The following preprocessing pipeline was applied to all original and augmented fMRI data: images were head-motion corrected through affine realignment of frames, brain-extracted as described in sect. \ref{subsec:augmentation} above, spatially normalized to the MNI152 EPI template, and smoothed with a 6 mm Gaussian kernel. Note that in contrast to the T1-based coregistration used during data augmentation, where the priority was to accurately warp a subject's brain into a different anatomy, a direct EPI-based registration was used for fMRI spatial normalization. Direct warping of individual fMRI images onto an EPI template has been shown to be more accurate for normalization to a template than cross-modal normalization as it accounts for magnetic inhomogeneities particular to EPI images \cite{Dohmatob.2018, Calhoun.2017}.
The preprocessed fMRI images were fitted to subject-level generalized linear models (GLMs) in SPM12. The design matrix for the GLMs was defined as described in Greenberg et al. \cite{Greenberg.2015}, with regressors for each of the 3 conditions in the reward processing task. The fitted GLM coefficients for these regressors were projected back into voxel space to yield 3 contrast maps, i.e. spatial maps of BOLD response to each task condition. A study-specific brain parcellation was created from resting-state fMRI from all subjects in the EMBARC dataset using the spectral clustering method developed by Craddock et al. \cite{Craddock.2012}. This brain parcellation was used to compute the mean regional contrast values from each of the 3 contrast maps. These mean regional values are the input features for predictive model training, explained in the next section. The granularity of this parcellation (number of regions-of-interest [ROIs]) was optimized during the model selection process, during which 100-, 200-, and 400-ROI parcellations were tested.

\paragraph{Neural Network Construction, Model Search, and Validation}
\label{subsec:model_search}
A feed-forward fully connected neural network was chosen as the predictive model. The model takes as input mean ROI values from each contrast map plus demographic and clinical covariates, and it predicts the 8-week change in the HAMD depression score. A loss function based on the coefficient of determination ($R^2$) was used: $L(\mathbf{y},\mathbf{\hat{y}}) = 100(1-f_{R^2}(\mathbf{y},\mathbf{\hat{y}}))$ where $\mathbf{y}, \mathbf{\hat{y}}$ are the true and predicted HAMD scores and $f_{R^2}(.)$ computes the coefficient of determination for the set of points with x and y coordinates given through its arguments. Random search, a popular model selection method, was performed to optimize model hyperparameters, including but not limited to: number of layers, neurons per layer, activation function, dropout rate, learning rate, and parcellation granularity. Three hundred model configurations were randomly chosen over a hyperparameter search space. To obtain an unbiased estimate of real-world performance, the models were evaluated using nested K-fold cross-validation with 5 inner folds and 3 outer folds. Within each outer fold, model performance was ranked by mean $R^2$ across the held-out partitions of the inner folds. The model with the highest $R^2$ was selected from each outer fold and test $R^2$ was measured on the held-out partition of the outer fold, not used for model training nor selection. The final model performance was the mean test performance over the 3 outer folds.

Two model searches were performed including one \textit{without} data augmentation (search \textit{N}) and one \textit{with} data augmentation (search \textit{A}). To ensure fair comparisons, the 300 searched models, their initial weights, and the cross-validation splits were identical between the two model searches. In search \textit{A}, augmentation was applied to the training partition of each fold to increase sample size by a factor of 5. Specifically, each source subject in the training partition was augmented to 5 target subjects of the same gender and age decile. Target subjects were selected from other treatment groups without subject overlap in the EMBARC dataset to ensure that no target subjects came from the held-out partition. Demographics and clinical measurements were copied over to the 5 new synthetic samples without augmentation, so that the augmentation would only be synthesizing new fMRI data of a different but realistic brain shape, whilst assuming the remainder of the synthesized subject was constant. 

\section{RESULTS}
\label{sec:results}

\begin{figure}[ht]
\begin{center}
\includegraphics[width=\textwidth]{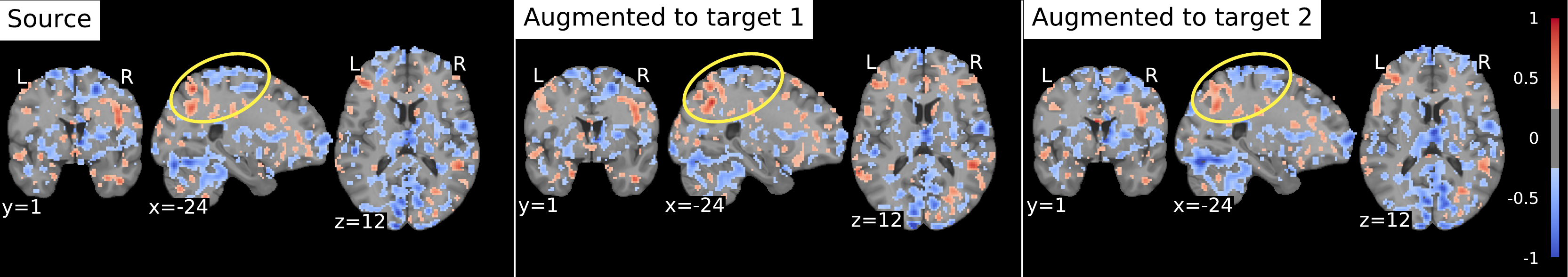}
\end{center}
\caption[example] 
{\label{fig:contrast_maps} 
Example fMRI GLM contrast maps derived from a source fMRI image and two augmented fMRI images synthesized from the source. A region with similar distributions of high and low intensity areas but distinct local variations is circled in yellow. Contrast map values were thresholded at 0.25 for clarity.}
\end{figure}

\paragraph{Data Augmentation}
Synthetized contrast maps are compared to the original source fMRI in \textbf{Fig. \ref{fig:contrast_maps}}. While all three contrast maps have similar high and low intensity areas globally across the brain, locally the shape and location of the individual clusters do vary (\textit{circled}) which confirms that the augmentation procedure generated data samples with distinct spatial variations. 

\paragraph{Model Search Results}
The top model from the non-augmented model search $N$, referred to here as $N_1$, achieved $R^2$ of 11.2\% and RMSE of 6.57 in predicting HAMD change. The top model from augmented search $A$, here called $A_1$, attained $R^2$ of 14.1\% and RMSE of 6.46 in the 52-point HAMD score. This constitutes a substantial 26\% relative increase in $R^2$. To ascertain the significance of this finding, these top models were retrained 100 times each with random weight initializations and $R^2$ and RMSE were compared with a two-tailed t-test. The differences in $R^2$ and RMSE of $N_1$ and $A_1$ were both significant with $p = 0.001$. Comparisons of these performance gains to those in the literature are described in the discussion section below.

\paragraph{Ablative Experiment}
To test the impact of data augmentation in model training, $A_1$ was retrained without augmented data to create $A_1'$. Without the augmented data, performance decreased from $R^2$ of 14.1\% and RMSE of 6.46 to $R^2$ of 10.7\% and RMSE of 6.99. As a further test, the top 5 models from each of the 3 outer folds of search $A$ ($A_1, A_2, ..., A_{15}$) were retrained without augmented data creating $A_1', A_2', ..., A_{15}'$. A pairwise two-tailed t-test between $A_i$ and $A_i'$ demonstrated \textit{a significant decrease in performance when augmented data was removed}: $R^2$ decreased by $5.8 \pm 5.1\%$ ($p = 0.0006$) and RMSE increased by $0.21 \pm 0.18$ ($p = 0.0006$).

\paragraph{Additive experiment}
In a reciprocal test of the impact of data augmentation in model selection, $N_1$ was retrained with augmented data. This new model, $N_1'$ did not exhibit increased performance with $R^2$ of 11.2\% and RMSE of 6.74. This comparison was extended to the top 5 models from each of the 3 outer folds of search $N$ ($N_1, N_2, ..., N_{15}$), which were retrained with augmented data creating $N_1', N_2', ..., N_{15}'$. A pairwise two-tailed t-test between $N_i$ and $N_i'$ revealed a non-significant improvement: $R^2$ increased by $1.5 \pm 4.4\%$ ($p = 0.209$) and RMSE decreased by $0.05 \pm 0.16$ ($p = 0.214$).

\section{DISCUSSION AND CONCLUSION}
\label{sec:discussion}
This work introduces a data augmentation method for synthesizing fMRI images with realistic brain morphology and demonstrates its performance benefit in a predictive task. The overall best performance in predicting antidepressant response was achieved by performing augmentation before model search. The best model using this approach, $A_1$, outperformed the best model from a search conducted without augmentation, $N_1$, by 26\% in $R^2$. This result compares favorably with the 12\% relative improvement achieved by state-of-the-art augmentation methods for natural images and the 7-8\% relative improvement achieved by a previous sMRI augmentation method \cite{Castro.41520154182015, Ulloa.91720159202015}. Additionally, this method was shown to provide the most benefit when used not only for model training, but also throughout the model search process. In the ablative experiment, models selected from search \textit{A} performed significantly worse when retrained without augmented data. In fact, the previous best model $A_1'$ performed worse than $N_1$, suggesting that the high-performing models found in search \textit{A} would have been missed by search \textit{N}. These observations strongly indicate that data augmentation should be performed prior to model hyperparameter optimization. Conversely, the additive experiment showed a lesser performance benefit when augmented data was introduced after the model search. This suggests that search \textit{N} identified less statistically powerful models that could not increase performance when data was augmented after the search.

While these results were limited to one task-based fMRI dataset, additional work will demonstrate generalization to additional datasets and resting-state fMRI. Another possible limitation arose from the T1-based coregistration employed in the augmentation. The source fMRI-to-source sMRI cross-modal registration may be imprecise due to EPI-specific non-linearities, causing the source fMRI to not be exactly registered to the target brain. Future work may test a direct EPI-based coregistration to the target fMRI. Finally, future experiments will test more extensive augmentation such as to 10-20x the original dataset size rather than 5x. Despite these limitations, the current findings show that the proposed fMRI augmentation can already significantly improve deep learning performance on neuroimaging predictive tasks.

In conclusion, this work proposes a novel, coregistration-based fMRI data augmentation method to synthesize realistic fMRI images that requires no new expensive fMRI acquisition. The method demonstrates improved performance in a challenging prediction task of antidepressant response prediction. This work also provides evidence that augmentation should precede hyperparameter optimization and that augmentation not only improves overall model performance but also the identification of better models during model hyperparameter optimization. We look forward to extending this promising approach to further increase its benefits. 

\bibliography{report} 
\bibliographystyle{spiebib} 
\end{document}